\documentclass{article} 
\usepackage{iclr2020_conference,times}

\usepackage{amsmath,amsfonts,bm}

\def\eqref#1{equation~\ref{#1}}

\def\1{\bm{1}}

\DeclareMathAlphabet{\mathsfit}{\encodingdefault}{\sfdefault}{m}{sl}
\SetMathAlphabet{\mathsfit}{bold}{\encodingdefault}{\sfdefault}{bx}{n}

\usepackage{hyperref}
\usepackage{url}
\usepackage{graphicx}
\usepackage{algorithm}
\usepackage{algpseudocode}
\usepackage{amsthm}      
\usepackage[normalem]{ulem}
\usepackage[nobottomtitles]{titlesec}

\newcommand{\dkl}[2]{\mathbb{D}_\text{KL}[#1 \vert \vert #2]}

\newcommand{\LL}[0]{\mathcal{L}}

\newcommand{\avg}[2]{\left\langle #1 \right\rangle_{#2}}

\newcommand{\pt}[1]{p_\theta \left(#1 \right)}

\newcommand{\qp}[1]{q_\phi \left(#1 \right)}
\newcommand{\intg}[2]{\int #1 \, \mathrm{d} #2}

\newcommand{\N}[1]{\mathcal{N} \left( #1 \right)}

\newcommand{\vect}[1]{\mathrm{vec} \left( #1 \right)}

\newtheorem*{remark}{Remark}
\newcommand{\defeq}{\mathrel{\mathop:}=}

\makeatletter
\newcommand*\bigcdot{\mathpalette\bigcdot@{.5}}
\newcommand*\bigcdot@[2]{\mathbin{\vcenter{\hbox{\scalebox{#2}{$\m@th#1\bullet$}}}}}
\makeatother

\title{Product Kanerva Machines:\\ Factorized Bayesian Memory}

\author{Adam H. Marblestone\textsuperscript{*}, Yan Wu\thanks{Equal contribution} \space\& Greg Wayne\\
Google DeepMind\\
London, N1C 4AG, UK \\
\texttt{\{amarbles,yanwu,gregwayne\}@google.com}}

\iclrfinalcopy 
\begin{document}

\maketitle

\begin{abstract}
An ideal cognitively-inspired memory system would compress and organize incoming items. The Kanerva Machine~\citep{dkm2018,wu2018kanerva} is a Bayesian model that naturally implements online memory compression. However, the organization of the Kanerva Machine is limited by its use of a single Gaussian random matrix for storage. Here we introduce the Product Kanerva Machine, which dynamically combines many smaller Kanerva Machines. Its hierarchical structure provides a principled way to abstract invariant features and gives scaling and capacity advantages over single Kanerva Machines. We show that it can exhibit unsupervised clustering, find sparse and combinatorial allocation patterns, and discover spatial tunings that approximately factorize simple images by object. 
\end{abstract}

\section{Introduction}
Neural networks may use external memories to flexibly store and access information, bind variables, and learn online without gradient-based parameter updates~\citep{fortunato2019generalization, graves2016hybrid, wayne2018unsupervised, sukhbaatar2015end, Banino2020MEMO, bartunov2019meta, munkhdalai2019metalearned}. Design principles for such memories are not yet fully understood.

The most common external memory is slot-based. It takes the form of a matrix $M$ with columns considered as individual slots. To read from a slot memory, we compute a vector of attention weights $\mathbf{w}$ across the slots, and the output is a linear combination $z_{\text{read}} = M \mathbf{w}$. Slot memory lacks key features of human memory. First, it does not automatically compress -- the same content can be written to multiple slots. Second, the memory does not naturally organize items according to relational structure, i.e., semantically related items may have unrelated addresses. Third, slot memory is not naturally generative, while human memory supports imagination~\citep{schacter2016remembering}. In addition, human memory performs novelty-based updating, semantic grouping and event segmentation~\citep{gershman2014statistical, franklin2019structured, howard2007semantic, koster2018big}. It also seems to extract regularities across memories to form ``semantic memory''~\citep{tulving1972episodic}, a process likely related to systems consolidation~\citep{kumaran2016learning}.

The Kanerva Machine ~\citep{dkm2018,wu2018kanerva, gregor2019shaping} replaces slot updates with Bayesian inference, and is naturally compressive and generative. Instead of a matrix $M$, it maintains a distribution $p(M)$. Reading with a vector of weights $\mathbf{w}$ over columns of $M$ -- which specify a query or address for lookup-- corresponds to computing a conditional probability of an observation $p(\mathbf{z}|M,\mathbf{w})$, while writing corresponds to computing the posterior given an observation $p(M|\mathbf{z}, \mathbf{w})$. The Kanerva Machine has a few disadvantages, due to its flat memory structure. First, computationally, it scales poorly with the number of columns $m$ of $M$, $\mathcal{O}(m^3)$ for inferring optimal weights $\mathbf{w}$. Second, it distributes information across all parameters of the memory distribution without natural grouping. 

To remedy both problems we introduce the hierarchically structured \emph{Product Kanerva Machine}. Instead of a single Kanerva Machine of $m$ columns, we divide the model into $k$ machines each with $m_i = m/k$ columns. Readouts from each of the $k$ machines are combined using weights $r_i$ inferred by an assignment network (Fig. 1A). Multi-component memory is inspired by neuroscience models of the gating of memory by contextual signals~\citep{Podlaski2020.01.08.898528, basu2016gating, pignatelli2019engram}. Factorizing a Kanerva Machine brings a computational speed advantage, and allows the individual machines within to specialize, leading to meaningful grouping of information.

\begin{figure}[h]
\begin{center}
\vspace*{-3mm}
\includegraphics[scale=0.69]{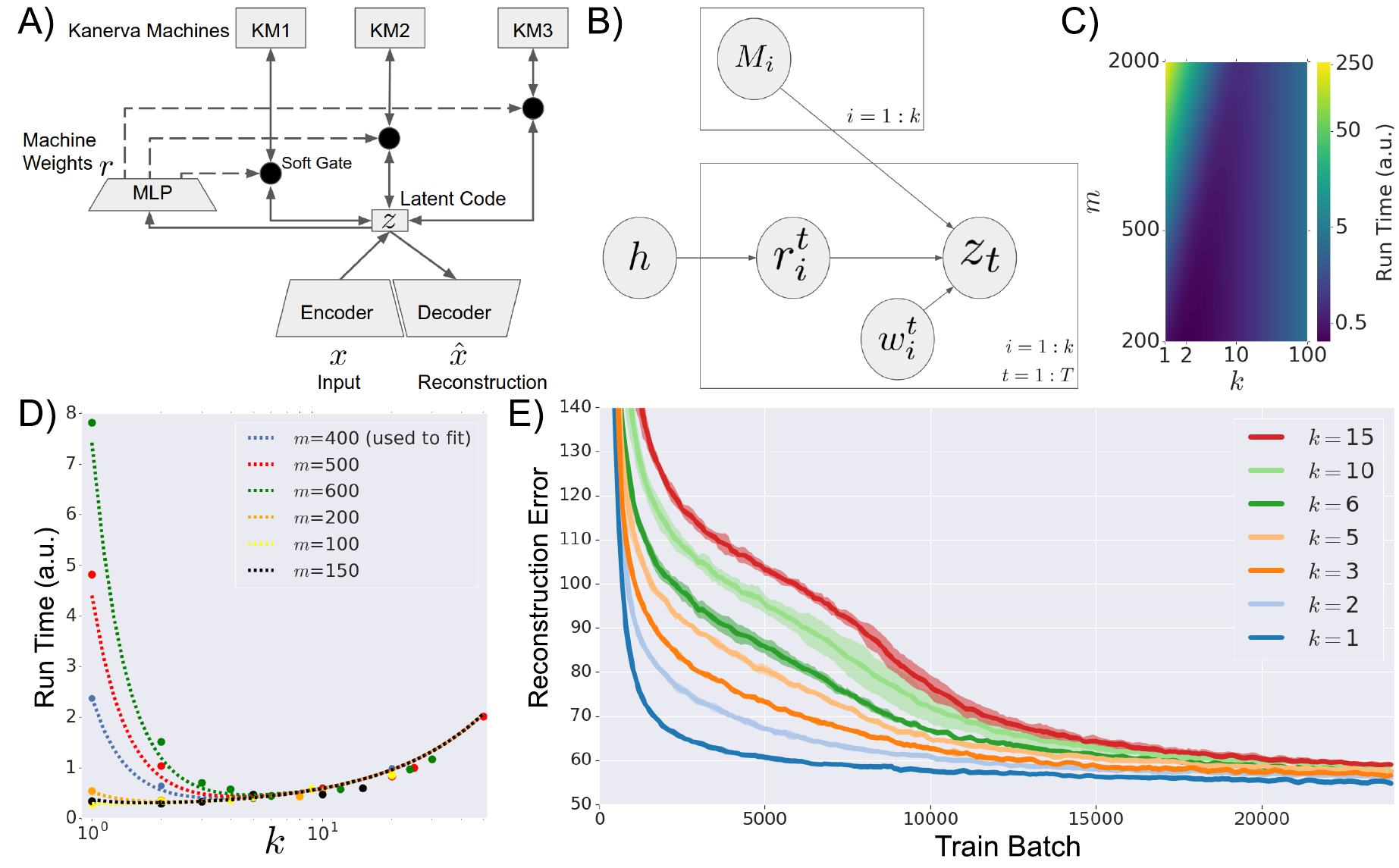}
\end{center}
\vspace*{-6mm}
\caption{Product Kanerva Machine concept and scaling. A) Architecture. B) Generative model. C) Theoretical scaling of run time with machines $k$ and total columns $m$.D) Fit to empirical scaling for $m=100:600$. E) Training curves for MNIST reconstruction at fixed $m=30$ and varying $k$.}
\end{figure}

\section{The Product Kanerva Machine}
The memory system (Fig. 1A) is composed from $k$ small Kanerva Machines each with $m_i$ columns and $c$ rows, where $c$ is the latent code size and $m_i = m/k$ is the number of memory columns per single Kanerva Machine. In our experiments, an encoder/decoder pair is used to map between images $\mathbf{x}$ and latent codes $\mathbf{z}$. An assignment network, here a simple multilayer perceptron (see Supp.~
\ref{trainingdetailsandhyperparameters} for details), is used to compute soft weights $\{r_i\}$ that define the relative strength of reading from or writing to the $i$th Kanerva Machine. The model supports writing, queried reconstruction (i.e., ``reading''), and generation. When generating (Fig. 1B), the assignment network is conditioned on a history variable $\mathbf{h}$, while when writing to the memory it is conditioned on the current $\mathbf{z}$, and when reading it is conditioned on the read query $\mathbf{z_\text{query}}$. Column weights $\mathbf{w_i}$ are optimized by least-squares for reconstructing the query $\mathbf{z_{\text{query}}}$  (see Supp.~\ref{trainingdetailsandhyperparameters} for details).

The $i$th Kanerva Machine has the matrix normal distribution $p(\vect{M_i}) \sim \N{\vect{R_i}, V_i \otimes I}$ where $R_i$ is a $c \times m_i$ matrix containing the mean of $M_i$, with $m_i$ its  number of columns, $V_i$ is a $m_i \times m_i$ matrix giving the covariance between columns, vectorization means concatenation of columns, and the identity matrix $I$ is $c \times c$. Given addressing weights $\mathbf{w_i}$ for machine $i$, the read-out from memory is the conditional distribution $p(\mathbf{z} | M_i) \sim \N{M_i \, \mathbf{w_i}, \sigma^2_i \bigcdot I}$. 

Two possible factorizations are mixtures and products. A product model assumes a factorized likelihood $p(\mathbf{z}) \propto \prod_{i=1}^k\, p(\mathbf{z}| M_i)$, which encourages each component $p(\mathbf{z}|M_i)$ to extract combinatorial, i.e., statistically independent, features across episodes~\citep{williams2002products, hinton1999products, hinton2002training, welling2007product}. A product factorization could therefore comprise a prior encouraging disentanglement across episodes, a milder condition than enforcing factorization across an entire dataset~\citep{locatello2019, burgess2019monet,higgins2018towards, watters2019spatial}. A mixture model (or the related switching models~\citep{NIPS2008_fox}), on the other hand, tends to find a nearest-neighbour mode dominated by one canonical prototype~\citep{hasselblad1966estimation,shazeer2017outrageously}. To address both scenarios, we use a ``generalized product''~\citep{cao2014generalized, peng2019mcp} (Eq.~\ref{gpmodel}), containing both products and mixtures as limits (see Supp.~\ref{mixturemodeldef}). We thus consider a joint distribution between $\mathbf{z}$ and all $k$ memory matrices, with each term raised to a power $r_i\geq0$
\begin{align}
\label{gpmodel}
    p(\mathbf{z}, M_1, \dots M_i, \dots) &\propto \prod_{i=1}^k \, p(\mathbf{z}, M_i)^{r_i}
\end{align}

During writing, $\{r_i\}$ are inferred from the observation $\mathbf{z}$, and a variable $\mathbf{h}$ which stores information about history. We use $p(\{r_i\}|\mathbf{h})$ during generation and an approximate posterior $q(\{r_i\}|\mathbf{z}, \mathbf{h})$ during inference (see Supp.~\ref{sec:inferencemodeldetail} for details).
Once $\{r_i\}$ are given, Eq. 1 becomes a product of Gaussians, which stays in the linear Gaussian family, allowing tractable inference~\citep{roweis1999unifying}. Given $\mathbf{z}$ and $\{r_i\}$, writing occurs via a Bayesian update to each memory distribution $p(M_i | \mathbf{z}, \{r_i\})$. Updates for the $k$ memories, given $\mathbf{z}$ are (see Supp.~\ref{productmodelderivation} for derivation)
\begin{align}
    \Delta &= \mathbf{z} - \mathbf{\mu_z}  \label{eq:prod-error} \\
    R_i &\leftarrow R_i + \beta_i \, \Delta \, \mathbf{w_i^\top} \, V_i \label{eq:prod-m-update} \\
    V_i &\leftarrow V_i - \beta_i \, V_i \, \mathbf{w_i}  \mathbf{w_i^\top} \, V_i
    \label{eq:prod-v-update}
\end{align}
where \begin{align*}\beta_i = \frac{1}{\mathbf{w_i^\top} \, V_i \, \mathbf{w_i} + \sigma^2_i / r_i} \text{  and  } \mathbf{\mu_z} = \frac{\sum_{i=1}^k \frac{r_i}{\sigma_i^2} \, R_i \, \mathbf{w_i}}{\Sigma_{j=1}^k \frac{r_j}{\sigma_j^2}} = \sum_{i=1}^k \gamma_i R_i \mathbf{w_i} \end{align*} 
For reading, $\mathbf{\mu_z}$ is used as the memory readout. Note how the prediction error term $\Delta$ (as in a Kalman Filter) now  \emph{couples} the $k$ machines, via $\mathbf{\mu_z}$. Algorithms for writing/reading are given in Supp.~\ref{algorithmdef}. The generative model (Fig. 1B) is trained by maximizing a variational lower bound~\citep{kingma2013auto} $\LL$ on $\ln\pt{\mathbf{x}}$ derived in Supp.~\ref{generativemodelderivation} (see Supp.~\ref{sup-condgen} for conditional generations).

\section{Results}
\label{results}

\subsection{Scaling}

We first asked if a product factorization could give a computational advantage. For a single Kanerva Machine, solving for $\mathbf{w}$ scales as $\mathcal{O}(m^3)$ due to the use of a Cholesky decomposition in the least-squares optimization. Parallel operation across $k$ machines gives theoretical scaling of $\mathcal{O}((\frac{m}{k})^3)$. If there are substantial fixed and per-machine overheads, we predict a scaling of the run time of $c + ak + b(m/k)^3$, with optimum at $k_\text{opt}=(\frac{3bm^3}{a})^\frac{1}{4}$. The empirically determined scaling of the run-time matches\footnote{Parameters $a = \text{3.318e-08} \pm \text{1.035e-09}$, 
$b = \text{2.176e-01} \pm \text{3.913e-02}$, $c = \text{3.676e-02} \pm \text{3.787e-03}$ fit to $m=400$ with $R^2 = 0.996$, which then explain $m=600$ with $R^2 = 0.992$ and $m=100$ with $R^2 = 0.958$.} this model (Fig. 1C-D). For $m>500$, a large speed advantage results even for moving from $k=1$ to $2$, showing computational benefit for product factorization.

\subsection{Queried reconstruction}
We began with a simple queried reconstruction task. A memory of $m=30$ total columns was divided into $k=1$ to $k=15$ machines. A set of 45 MNIST digits were written, and then the memory was read using each item as a query, and the average reconstruction loss $\avg{\ln \pt{\mathbf{x} | \mathbf{z}}}{\qp{\mathbf{z}}}$ was computed. All factorizations eventually achieved similar reconstruction accuracy (Fig. 1E), showing that product factorization does not incur a loss in representational power.

\subsection{Pattern completion}
\begin{figure}[h]
\begin{center}
\vspace*{-4mm}
\includegraphics[scale=0.54]{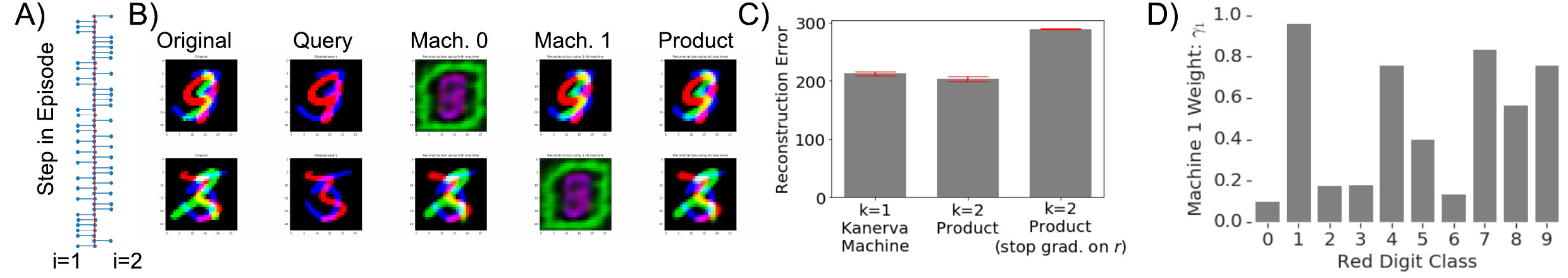}
\end{center}
\vspace*{-6mm}
\caption{RGB binding task demonstrating sparse machine usage and unsupervised classification. A) For each step in an episode, machine weights $\gamma_0$, $\gamma_1$ are displayed as a thorn plot, showing sparsity of usage. B) From left to right: original image, query image, reconstructions from each machine (with other machine blanked) given query image, reconstruction from full product model given query image. C) Performance when factorizing $m=60$ total columns into $k=2$ machines of size $30$ vs. a single Kanerva Machine (and stop gradient to assignment weights $r_i$ demonstrating that $\mathbf{r}$ is optimized). D) Unsupervised partial MNIST digit classification by the model with $k=2$: the machine assignment weight $\gamma_1$ is highly correlated with the Red digit class.}
\end{figure}

We next tested the Product Kanerva Machine on the storage of associations/bindings between high-dimensional variables, to ask if a product ($k=2$) model might show an advantage over a single Kanerva Machine ($k=1$). A set of 45 triplets of MNIST digits were stored in memories with $m=60$ total columns, each triplet consisting of an MNIST digit for the Red, Green and Blue channels of an image. Partial queries consisting of the R and B channels, but not the G channel, were presented and the average reconstruction loss was computed across all 3 channels.

For $k=2$ machines, the system finds a sparse machine usage pattern (Fig. 2A), with $r_i\approx0$ or $1$, but both machines used equally overall. When a given individual machine is used, it reconstructs the full bound pattern, while the unused machine produces a fixed degenerate pattern (Fig. 2B). A product of $k=2$ machines each of 30 columns outperforms a single machine of 60 columns, while a stop gradient on $r$ abolishes this (Fig. 2C). Choice of $r$ depends on digit class (Fig. 2D), e.g., in Fig. 2D predominantly on the R digit but not the B or G digits (symmetry is broken between B and R from run to run leading to horizontal or vertical stripes) -- see Supp.~\ref{allrgbclasses} for full RGB class selectivity matrix. Thus, the model optimizes allocation via sparse, dynamic machine choice which becomes selective to digit class in an unsupervised fashion.

\subsection{Within-item factorization}
\begin{figure}[h]
\begin{center}
\vspace*{-4mm}
\includegraphics[scale=0.54]{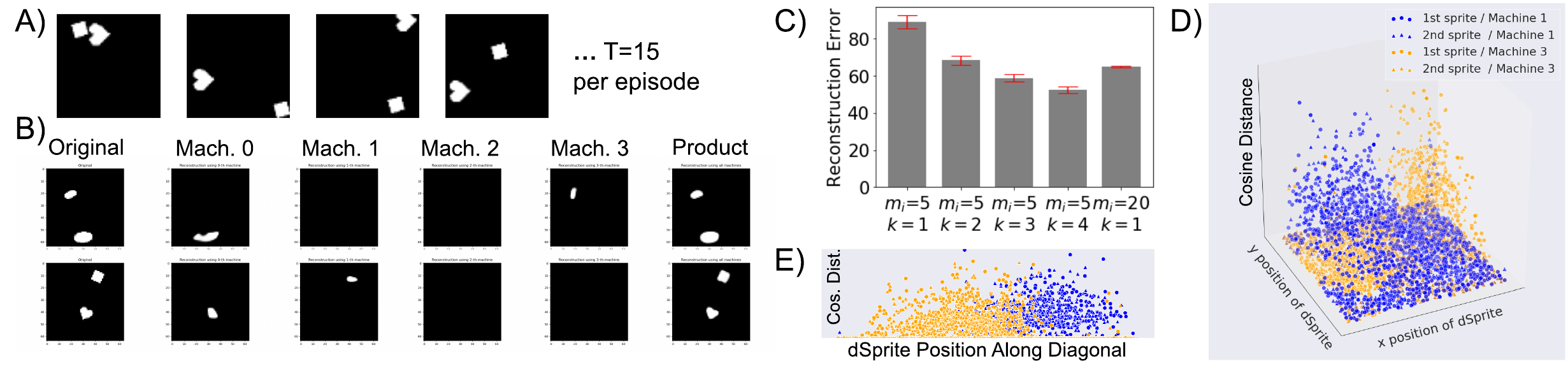}
\end{center}
\vspace*{-6mm}
\caption{``Dancing dSprites'' task, demonstrating item factorization across multiple machines. A) Task design. B) Factorization examples. C) Reconstruction loss vs. $k$ and $m_i$. D-E) Spatial tunings of individual machines: cosine distance between the reconstruction from each machine and template dSprite vs. template position. Diagonal in E spans from lower right to upper left in D.}
\end{figure}

We next asked whether multiple pieces of content extracted from within \emph{single} items could be differentially routed. To probe the factorization of multiple ``objects'' across machines, we developed a simple ``dancing dSprites'' task (Fig. 3A). In any episode, 15 images were written, each with the same combination of two dSprites~\citep{dsprites17}, in randomized positions from image to image within an episode, and where shapes, orientations and scales varied across episodes. 

For $k=4$ machines, each with 5 columns, we observed a form of object-based factorization (Fig. 3B): \emph{individual} machines typically reconstructed distorted images at the positions of single objects (more examples in Supp.~\ref{sup-objectfactor}). The $k=4$ model outperformed a $k=1$ model with the same total number of columns (Fig. 3C). Individual machine reconstructions exhibited localized spatial tunings to the positions of the individual dSprites (Fig. 3D,E and Supp.~\ref{supp:otherinvariances}). In contrast, tuning was invariant to the shape, orientation and size of the dSprites (Supp.~\ref{supp:otherinvariances}). Weights $r_i$ were nearly fixed, suggesting that selectivity was not due to varying $r$. The model thus spontaneously factored according to localized spatial tunings, such that single machines typically reconstructed single objects. 

\section{Future directions}
Product Kanerva Machines could be extended in several ways. Attention-based selection of input elements could be added, or explicit event segmentation over time, or alternative gating methods and forms of communication between machines, as in~\citep{goyal2019recurrent, santoro2018relational, hinton2018matrix, veness2017online,kipf2018compile}. Auxiliary losses could encourage richer unsupervised classification~\citep{makhzani2015adversarial} for class-dependent routing. The generative model can be extended with richer distribution families~\citep{rezende2015variational}. Joint inference of $\mathbf{r}$ and $\mathbf{w}_i$ using Expectation Maximization (EM) algorithms may be possible~\citep{dempster1977maximum}. Further understanding of when and how factorized memories can encourage extraction of objects or other disentangled features may also be of interest. Ultimately, we hope to use compressive, semantically self-organizing and consolidating memories to solve problems of long-term credit assignment~\citep{ke2018sparse, hung2019optimizing}, continual learning~\citep{van2018generative, rolnick2019experience} and transfer~\citep{higgins2017darla}. 

\subsubsection*{Acknowledgments}
We thank Andrea Banino, Charles Blundell,  Matt Botvinick, Marta Garnelo, Timothy Lillicrap, Jason Ramapuram, Murray Shanahan and Chen Yan for discussions, Sergey Bartunov for initial review of the manuscript, Loic Matthey, Chris Burgess and Rishabh Kabra for help with dSprites, and Seb Noury for assistance with speed profiling.

\bibliography{references}
\bibliographystyle{iclr2020_conference}

\appendix
\section{Supplemental Materials}

\subsection{Experimental details and hyper-parameters}
\label{trainingdetailsandhyperparameters}

\subsubsection{Hyper-parameters}
The model was trained using the Adam optimizer~\citep{kingma2014adam} with learning rate between $5e^{-5}$ and $1e^{-3}$ and batch size $24$. 

For the RGB binding task, a high learning rate of $1e^{-3}$ was used and encouraged fast convergence to the sparse machine weights solution. For learning rate of $1e^{-4}$ or below, the $k=2$ model initially underperformed the $k=1$ model with the same total columns but then rapidly switched to the sparse solution and superior performance after roughly 200000 train batches.

Latent code sizes $c$ were typically $50$ but were $100$ for the RGB binding task. The size of the history variable $\mathbf{h}$ was $10$.

Convolutional encoders/decoders with ReLU activations were used 
\begin{itemize}
\item Encoder: output channels [16, 32, 64, 128], kernel shapes (6, 6), strides 1 
\item Decoder: output channels [32, 16, 1 for grey-scale or 3 for RGB images], \\output shapes [(7, 7), (14, 14), (28, 28)], kernel shapes (4, 4), strides 2
\end{itemize}
except in the case of the dancing dSprites task where a small ResNet (2 layers of ResNet blocks with leaky ReLU activations each containing 2 convolutional layers with Kernel size 3, with an encoder output size 128 projected to $c$ and using pool size 3 and stride 2) was used in order to improve reconstruction quality for dSprites.

\subsubsection{Treatment of addressing weights $\mathbf{w}$}
For solving for the least-squares optimal read weights $\mathbf{w}$, we used the matrix solver $matrix\_solve\_ls$ in TensorFlow, in Fast mode with L2 regularizer $\lambda=0.1$ to $0.5$, typically $0.35$.

\subsubsection{Treatment of machine assignment weights $r$}
Logits for choosing the $k$ machine weights $\{r_i\}$, in $q_\phi(r^t_{1:k} | \mathbf{h}_{t-1}, \mathbf{z}_t)$ or $P_\theta (r^t_{1:k}|\mathbf{h}_{t-1})$, parametrizing a diagonal Gaussian $\N{\mathbf{\mu}_r, \mathbf{\sigma}_r^2}$ in the $\ln (r)$ space (see Supp.~\ref{generativemodelderivation}), were created as follows. During reading and writing, we used $\mathbf{\mu_r} = \text{Linear}_1([\text{MLP}_1(\mathbf{z_t}), \mathbf{h}_{t-1}])$, $\mathbf{\sigma_r} = \text{Linear}_2([\text{MLP}_1(\mathbf{z_t}), \mathbf{h}_{t-1}])$. During generation, we used $\mathbf{\mu_r} = \text{Linear}_3(\mathbf{h})$, $\mathbf{\sigma_r} = \text{Linear}_4(\mathbf{h})$. $\text{MLP}_1$ had layer widths [40, 20, $k$]. Samples from the resulting Gaussian were passed through a SoftPlus function to generate effective machine observation noises~\citep{wu2018kanerva} $\sigma_i/\sqrt{r_i}$, and then squared, inverted and normalized to generate the overall machine weight $\gamma_i = \frac{r_i}{\sigma_i^2} / \sum_{j=1}^k \frac{r_j}{\sigma_j^2}$. See Supp.~\ref{algorithmdef} for full machine choice algorithm and Supp.~\ref{generativemodelderivation} for definitions of the distributions in the generative and inference models.

\subsubsection{Speed tests}
Speed tests were performed on a V100 GPU machine with 8 CPU cores, with memory operations assigned to CPU to encourage parallelization and encoder/decoder operations assigned to the GPU. 

\subsubsection{Tuning analysis}
To analyze the tunings of dSprite reconstruction to dSprite properties (Fig. 3D-E), we used a template matching procedure. A template image with individual dSprites at their original individual positions in the stored images was matched to each machine's reconstruction $\hat{x}$ via a cosine distance on the image pixel vector $\frac{\hat{x}_\text{machine i} \bigcdot \text{dSprite}_j}{\|\hat{x}_\text{machine i}\|\|\text{dSprite}_j\|}$ where $i$ indexes over the $k$ machines and $j$ indexes over the two dSprites in each image.

\subsection{Generative model definition}
\label{generativemodelderivation}

The generative model (Fig. 1B, see Supp.~\ref{sup-condgen} for example conditional generations) is trained by maximizing a variational lower bound~\citep{kingma2013auto} $\LL$ on $\ln\pt{\mathbf{x}}$. For fixed machine weights $r_i$, we would use an ELBO
\begin{equation}
\LL = \avg{\ln \pt{\mathbf{x} | \mathbf{z}}}{\qp{\mathbf{z}}} - \sum_{i=1}^k \left[ \dkl{\qp{\mathbf{w_i}}}{\pt{\mathbf{w_i}}} + \dkl{\qp{M_i}}{\pt{M_i}} \right]
\label{eq:elbo}
\end{equation}
where $\pt{\mathbf{w_i}} = \N{0, 1}$. Here, we further consider the exponential weightings $r_i$ in the generalized product to depend on a latent variable $\mathbf{h}$ that summarizes the history, via $p(r_i | \mathbf{h})$. This gives a joint distribution
\begin{align}
    p(\mathbf{z}, \{M_i\}^k_{i=1}, \{r_i\}^k_{i=1}, \mathbf{h}) = p(\mathbf{z}, \{M_i\}^k_{i=1} | \{r_i\}^k_{i=1}) \, \prod_{i=1}^k p(r_i | \mathbf{h}) \, p(\mathbf{h})
\end{align}
where $p(\mathbf{z}, \{M_i\}^k_{i=1} | \{r_i\}^k_{i=1}) \propto \prod_{i=1}^k \, p(\mathbf{z}, M_i)^{r_i}$ and results in additional KL divergence terms in the ELBO
\begin{align}
    - \sum_{i=1}^k\dkl{q(r_i)}{p(r_i|\mathbf{h})} - \dkl{q(\mathbf{h})}{p(\mathbf{h})}
\end{align}

The full joint distribution is
\begin{align}
     P_\theta (\mathbf{z}_{1:T}, M^{1:T}_{1:k}, \mathbf{w}^{1:T}_{1:k}, r^{1:T}_{1:k}, \mathbf{h}) = P_\theta(M_{1:k}) \prod_{t=1}^T \left(P_\theta (\mathbf{z_t} | M^t_{1:k}, \mathbf{w}^t_{1:k}, r^t_{1:k}) P_\theta(r^t_{1:k}|h) P_\theta(\mathbf{w}^t_{1:k})\right) P_\theta(\mathbf{h}) \\
     = \prod_{t=1}^T \prod_{i=1}^k P_\theta(M^t_i) \prod_{t=1}^T \left(P_\theta (\mathbf{z_t} | M^t_{1:k}, \mathbf{w}^t_{1:k}, r^t_{1:k}) P_\theta(r^t_{1:k}|\mathbf{h}_{t-1}) \prod_{i=1}^k P_\theta(\mathbf{w}^t_i)\right) \prod_{t=1}^T P_\theta(\mathbf{h}_{t-1})
     \label{eq:generative-full-joint}
\end{align}

Marginalizing out $\mathbf{z}_{1:T}$, we have
\begin{align}
    P_\theta (M^{1:T}_{1:k}, \mathbf{w}^{1:T}_{1:k}, r^{1:T}_{1:k}, \mathbf{h})= \prod_{t=1}^T \prod_{i=1}^k P_\theta (M^t_i) \prod_{t=1}^T \prod_{i=1}^k P_\theta (\mathbf{w}^t_i) \prod_{t=1}^T P_\theta (r^t_{1:k}|\mathbf{h}_{t-1}) \prod_{t=1}^T P_\theta (\mathbf{h}_{t-1})
    \label{eq:generative-marginal}
\end{align}

$P_\theta (r^t_{1:k}|\mathbf{h}_{t-1})$ is a top down generative model for the grouping across machines of content from the individual component machines of the product model. In order to be able to generate sequentially, it must only depend on the history up to but not including the outputs from the present timestep, i.e., $\mathbf{h}_{t-1}$. Rather than parameterizing $\mathbf{r}$ as a distribution over categorical distributions, we instead parameterize $\ln (\mathbf{r})$ as a Gaussian (with trainable mean and diagonal variances), and then use a deterministic trainable network to produce $\mathbf{r}$.

$P_\theta (\mathbf{w}^t_i)$ is a standard Gaussian prior. 

$P_\theta (\mathbf{h}_{t-1})$ is a standard Gaussian prior. 

(As an alternative prior on $\mathbf{h}$, we can use a time-varying AR(1) process as the prior, $P_\theta(\mathbf{h}_{t-1} | \mathbf{h}_{t-2})$: this will allow the history variable to perform a random walk within an episode while slowly decaying to a standard Gaussian over time, as was used in~\cite{merel2018neural}.)

$P_\theta (M^t_i)$ is the trainable matrix Gaussian prior of each Kanerva Machine in the product model.

$P_\theta (\mathbf{z_t} | M^t_{1:k}, \mathbf{w}^t_{1:k}, r^t_{1:k})$ is the generation procedure for one step of the Product Kanerva Machine model as described elsewhere in this document. It will output the mean $\mathbf{w_i^\top} M_i$ from each machine and then combine them using the machine weights $\mathbf{r}$.

Note: we do not use any \emph{additional} prior $P_\theta(\mathbf{z})$, such as a standard Gaussian, and likewise when we have an encoder from the image to the latent $e(\mathbf{z}|\mathbf{x})$ we do not subject it to a VAE-style standard Gaussian prior, instead just using a simple autoencoder with conv-net encoder $e(\mathbf{z}|\mathbf{x})$ and deconv-net decoder $d(\mathbf{x}|\mathbf{z})$ outputting the $p$ parameter of a Bernoulli distribution for each image pixel.

\subsubsection{Inference model}
\label{sec:inferencemodeldetail}

We use the following factorization of the approximate posterior to infer the hidden variables given a sequence of observations

\small
\begin{align}
     q_\phi(M^{1:T}_{1:k}, \mathbf{w}^{1:T}_{1:k}, r^{1:T}_{1:k}, \mathbf{h} | \mathbf{z}_{1:T}) = q_\phi(M^{1:T}_{1:k} | \mathbf{w}^{1:T}_{1:k}, r^{1:T}_{1:k}, \mathbf{h}, \mathbf{z}_{1:T}) q_\phi(r^{1:T}_{1:k} | \mathbf{h}, \mathbf{z}_{1:T}) q_\phi(\mathbf{w}^{1:T}_{1:k} | z_{1:T}) q_\phi(\mathbf{h} | \mathbf{z}_{1:T}) \\= \prod_{t=1}^T q_\phi(M^t_{1:k} | \mathbf{w}^t_{1:k}, r^t_{1:k}, \mathbf{h}, \mathbf{z_t}, M^{t-1}_{1:k})
     \prod_{t=1}^T q_\phi(\mathbf{w}^t_{1:k} | \mathbf{z_t}, r^t_{1:k}, M^{t-1}_{1:k}) \prod_{t=1}^T q_\phi(r^t_{1:k} | \mathbf{h}_{t-1}, \mathbf{z}_t) \prod_{t=1}^T q_\phi(\mathbf{h}_{t-1} | \mathbf{z}_{1:t-1})
     \label{eq:inference-model}
\end{align}
\normalsize

$q_\phi(M^t_{1:k} | \mathbf{w}^t_{1:k}, r^t_{1:k}, \mathbf{h}, \mathbf{z}_t, M^{t-1}_{1:k})$ is the write step of our Product Kanerva Machine and is described elsewhere in this document

$q_\phi(\mathbf{w}^t_{1:k} | \mathbf{z}_t, r^t_{1:k}, M^{t-1}_{1:k})$ is the ``solve for $\mathbf{w}$ given query'' step of our Product Kanerva Machine and is performed by least-squares optimization. 

$q_\phi(r^t_{1:k} | \mathbf{h}_{t-1}, \mathbf{z}_t)$ is a bottom-up inference model producing the machine weights variable $\mathbf{r}$. Rather than parameterizing $\mathbf{r}$, we instead parameterize $\ln (\mathbf{r})$ as a Gaussian (with trainable mean and diagonal variances), and then use a deterministic trainable network to produce $\mathbf{r}$.

$q_\phi(\mathbf{h}_{t-1} | \mathbf{z}_{1:t-1})$ is where we will use a superposition memory to store a record of previous $\vec{z}$ and their associated $\mathbf{r}$ variables which will be used to produce the history variable $\mathbf{h}$. The superposition buffer takes the form
$\Omega_t = \frac{1}{t}\Psi([\mathbf{z_t}, r^t_{1:k}]) + \frac{t-1}{t}\Omega_{t-1}$
where $\Psi$ is a trainable embedding function. Then the distribution over the history variable $\mathbf{h}$ can be a diagonal Gaussian $q_\phi(\mathbf{h}_{t-1} | z_{1:t-1}) = N(\mathbf{\mu}, \mathbf{\sigma})$ where $\mathbf{\mu} = \text{MLP}_a(\Omega_{t-1})$ and $\mathbf{\sigma} = \text{MLP}_b(\Omega_{t-1})$ (we used small MLPs with layer widths $[10,10]$ here for $\text{MLP}_a$ and $\text{MLP}_b$). 

\bigskip
\bigskip
\bigskip
\bigskip
\bigskip
\bigskip
\bigskip
\pagebreak

\subsubsection{ELBO}

The full ELBO is

\begin{align}
& \ln P_\theta (\mathbf{z}_{1:T}) \geq \left< \frac{P_\theta (\mathbf{z}_{1:T}, M^{1:T}_{1:k}, \mathbf{w}^{1:T}_{1:k}, r^{1:T}_{1:k}, \mathbf{h})}{q_\phi(M^{1:T}_{1:k}, w^{1:T}_{1:k}, r^{1:T}_{1:k}, \mathbf{h} | \mathbf{z}_{1:T})} \right>_{q_\phi(M^{1:T}_{1:k}, \mathbf{w}^{1:T}_{1:k}, r^{1:T}_{1:k}, h | z_{1:T})}\\
& =\left< \ln P_\theta(\mathbf{z}_{1:T} | M^{1:T}_{1:k}, \mathbf{w}^{1:T}_{1:k}, r^{1:T}_{1:k}, \mathbf{h}) \right>_{q_\phi(M^{1:T}_{1:k}, \mathbf{w}^{1:T}_{1:k}, r^{1:T}_{1:k}, \mathbf{h} | z_{1:T})} \\ &- \dkl{q_\phi(M^{1:T}_{1:k}, w^{1:T}_{1:k}, r^{1:T}_{1:k}, \mathbf{h} | \mathbf{z}_{1:T})}{P_\theta(M^{1:T}_{1:k}, \mathbf{w}^{1:T}_{1:k}, r^{1:T}_{1:k}, \mathbf{h})} \\&=
\left< \ln P_\theta(\mathbf{z}_{1:T} | M^{1:T}_{1:k}, \mathbf{w}^{1:T}_{1:k}, r^{1:T}_{1:k}, \mathbf{h}) \right>_{q_\phi(M^{1:T}_{1:k}, \mathbf{w}^{1:T}_{1:k}, r^{1:T}_{1:k}, \mathbf{h} | \mathbf{z}_{1:T})} \\&-\dkl{\prod_{t=1}^T q_\phi(M^t_{1:k} | \mathbf{w}^t_{1:k}, r^t_{1:k}, \mathbf{h}, \mathbf{z}_t, M^{t-1}_{1:k})}{\prod_{t=1}^T \prod_{i=1}^k P_\theta (M^t_i)} \\&-\dkl{\prod_{t=1}^Tq_\phi(\mathbf{w}^t_{1:k} | \mathbf{z}_t, r^t_{1:k}, M^{t-1}_{1:k})}{\prod_{t=1}^T \prod_{i=1}^k P_\theta (\mathbf{w}^t_i)}\\&-\dkl{\prod_{t=1}^Tq_\phi(r^t_{1:k} | \mathbf{h}_{t-1}, \mathbf{z}_t)}{\prod_{t=1}^T P_\theta (r^t_{1:k}|\mathbf{h}_{t-1})} \\&- \dkl{
       \prod_{t=1}^T q_\phi(\mathbf{h}_{t-1} | \mathbf{z}_{1:t-1})}{  \prod_{t=1}^T P_\theta (\mathbf{h}_{t-1})}\\&=\left< \ln P_\theta(\mathbf{z}_{1:T} | M^{1:T}_{1:k}, \mathbf{w}^{1:T}_{1:k}, r^{1:T}_{1:k}, \mathbf{h}) \right>_{q_\phi(M^{1:T}_{1:k}, \mathbf{w}^{1:T}_{1:k}, r^{1:T}_{1:k}, \mathbf{h} | \mathbf{z}_{1:T})}\\&-
       \sum\limits_{t=1}^T \sum\limits_{i=1}^k \dkl{ q_\phi(M^t_{i} | \mathbf{w}^t_{i}, r^t_{i}, \mathbf{h}, \mathbf{z}_t, M^{t-1}_{i})}{ P_\theta (M^t_i)} \\&-\sum\limits_{t=1}^T \sum\limits_{i=1}^k \dkl{ q_\phi(\mathbf{w}^t_{i} | \mathbf{z}_t, r^t_{i}, M^{t-1}_{i})}{ P_\theta (\mathbf{w}^t_i)}\\&-\sum\limits_{t=1}^T\dkl{ q_\phi(r^t_{1:k} | \mathbf{h}_{t-1}, \mathbf{z}_t)}{ P_\theta (r^t_{1:k}|\mathbf{h}_{t-1})} \\&- \sum\limits_{t=1}^T \dkl{ q_\phi(\mathbf{h}_{t-1} | \mathbf{z}_{1:t-1})}{ P_\theta (\mathbf{h}_{t-1})}
     \label{eq:generative-product-elbo}
\end{align}

Regarding the term $\dkl{q_\phi(r^t_{1:k} | \mathbf{h}_{t-1}, \mathbf{z}_t)}{P_\theta (r^t_{1:k}|\mathbf{h}_{t-1})}$, this should ideally be a KL between two Dirichlet distributions~\citep{joo2019dirichlet}, i.e., between distributions over categorical distributions. Rather than parameterizing $\mathbf{r}$, we instead parameterize $\ln (\mathbf{r})$ as a Gaussian (with trainable mean and diagonal variances), and then use a deterministic trainable network to produce $\mathbf{r}$ itself. We are then left with Gaussian KLs which are easy to evaluate and Gaussian variables which are easy to re-parametrize in training.

\emph{Note}: For Fig. 1 and Fig. 2, we relaxed distributional constraints on $\mathbf{w}$ in the loss function in order to lower variance, by removing the KL loss $\dkl{ q_\phi(\mathbf{w}^t_{i} | \mathbf{z}_t, r^t_{i}, M^{t-1}_{i})}{ P_\theta (\mathbf{w}^t_i)}$ on $\mathbf{w}$ in the writing step (but not the reading step), and by using the mean $\mathbf{w}$ rather than sampling it. The full model was used in Fig. 3. The mixture model of Supp.~\ref{mixturemodeldef} was trained without the KL penalty $\dkl{ q_\phi(M^t_{i} | \mathbf{w}^t_{i}, r^t_{i}, \mathbf{h}, \mathbf{z}_t, M^{t-1}_{i})}{ P_\theta (M^t_i)}$ on $M$ to reduce variance, and also did not include KL terms for $\mathbf{h}$ or $\mathbf{r}$.

\subsubsection{Sampling from the generative model}
\label{howtogenerate}

To generate full episodes autoregressively: 
\begin{itemize}
\item We first sample priors $P_\theta (M_i)$, $P_\theta (\mathbf{w}_i)$ and $P_\theta(\mathbf{h})$ and then sample $P_\theta(r^t_{1:k}|\mathbf{h}_{t-1})$ and $P_\theta (\mathbf{z_t} | M^t_{1:k}, \mathbf{w}^t_{1:k}, r^t_{1:k})$ to produce $r^1_{1:k}$ and then $\mathbf{z}_1$. 
\item $\mathbf{z}_1$ is then decoded to an image $\mathbf{\hat{x}_1} = d(\mathbf{z}_1)$, each pixel of the image rounded to $0$/$1$ and then the image re-encoded as $e(\mathbf{\hat{x}_1})$. We then query the memory with the re-encoded image $e(\mathbf{\hat{x}_1})$ to obtain an updated $\mathbf{z}_1$. This step is repeated several times, 12 times here, to allow the memory to ``settle'' into one of its stored attractor states~\citep{dkm2018}. 
\item We then write $z_1$ into the product memory $M$ using the analytical memory update $q_\phi(M^t_{1:k} | \mathbf{w}^t_{1:k}, r^t_{1:k}, \mathbf{h}, \mathbf{z}_t, M^{t-1}_{1:k})$  of Eqs. 2-4 and $[\mathbf{z}_1,r^1_{1:k}]$ into the history $\mathbf{h}$ via $\Omega$ by using $\Omega_t \gets \frac{1}{t}\Psi([\mathbf{z_t}, r^t_{1:k}]) + \frac{t-1}{t}\Omega_{t-1}$. 
\item We then sample $q_\phi(\mathbf{h}_{t-1} | \mathbf{z}_{1:t-1})$ to produce $\mathbf{h}_1$ and sample $P_\theta(r^t|\mathbf{h}_{t-1})$ to produce $r^2$. 
\item We then read the memory, using as read weights a draw from the priors on $\mathbf{w}_i$, $P_\theta(\mathbf{w}_i)$, and as machine weights our $r^2$, which allows us to produce $\mathbf{z}_2$.  
\item ...and so on, until finished generating.
\end{itemize}

Note that if a partial episode has been written to begin with, we will simply have $M$ and $\Omega$ and hence $\mathbf{h}$ pre-initialized before starting this process rather than using their priors.

\subsection{Derivation of Product Kanerva write and read operations}
\label{productmodelderivation}

\subsubsection{Review of Kanerva Machine}

To derive the Product Kanerva Machine, we first reformulate a single Kanerva Machine in terms of a precision matrix rather than covariance matrix representation.

For a single Kanerva Machine, recall that the posterior update of the memory distribution $P(M_i | \mathbf{z})$ is given by the Kalman filter-like form
\begin{equation}
    \begin{split}
        R_i
        \leftarrow R_i +  \, (\mathbf{z} - R_i \, \mathbf{w_i}) \frac{1}{\mathbf{w_i^\top} \, V_i \, \mathbf{w_i} + \sigma_i^2} \mathbf{w_i^\top} V_i
    \end{split}
    \label{eq:i-mean}
\end{equation}
\begin{equation}
    V_i \leftarrow V_i - V_i \, w_i \, \frac{1}{\mathbf{w_i^\top} \, V_i \, w_i + \sigma_i^2}\,\mathbf{w_i^\top} \, V_i
    \label{eq:i-sigma}
\end{equation}

In addition, recall that we can analytically compute the mean and covariance of the joint distribution of $z$ and $M_i$, as well as of the marginal distribution of $z$ (integrating out $M_i$):
\begin{align}
    p(\mathbf{z}, M_i) = p(\mathbf{z} | M_i) \, p(M_i) \sim \N{\mu_i, \Sigma_i}
\end{align}
\begin{equation}
    p_i(\mathbf{z}) = \intg{p(\mathbf{z}|M_i)\,p(M_i)}{M_i} \sim \N{R_i\,\mathbf{w_i}, \, \underbrace{(\mathbf{w_i^\top} \, V_i \, \mathbf{w_i} + \sigma_i^2)}_{\Sigma_{z_i}} \bigcdot I}
\end{equation}
where
\begin{align}
    \mu_i &= \left[ \begin{matrix}
    R_i \, \mathbf{w_i} \\ \vect{R_i}
    \end{matrix} \right] \\
    \Sigma_i &= \left[ \begin{matrix}
    \Sigma_{z_i} \, \Sigma_{c_i}^\top \\ \Sigma_{c_i}, V_i
    \end{matrix} \right] \otimes I
\end{align}

The joint covariance is a Kronecker product of a block matrix, where the upper left block is $1 \times 1$ (a scalar), the upper right is $1 \times m_i$, the lower left is $m_i \times 1$ and the lower right is $m_i \times m_i$, and $I$ is the $c \times c$ identity matrix. 

To convert to the precision matrix representation, we can use the block matrix inversion rule to obtain the precision matrix for a single Kanerva Machine, similar to eqns. (10) and (11) in \cite{williams2002products}:
\begin{equation}
    \Lambda_i = \Sigma_i^{-1} = \left[ \begin{matrix}
    \sigma_i^{-2} & \Lambda^\top_{c_i} \\
    \Lambda_{c_i}& \Lambda_{M_i}
    \end{matrix} \right] \otimes I
\end{equation}
where 
\begin{align}
    \Lambda_{c_i} &= -\sigma^{-2}_i \, \mathbf{w_i} \\
    \Lambda_{M_i} &= V_i^{-1} + \mathbf{w_i}\,\sigma_i^{-2}\, w_i^\top \\
    &= \left(V_i - V_i \, \mathbf{w_i} \, (\sigma_i^2 + \mathbf{w_i^\top} \, V_i \, \mathbf{w_i})^{-1}\,\mathbf{w_i^\top} \, V_i \right)^{-1}\\
\end{align}
with the last step due to the Woodbury identity~\citep{bishop2006pattern}. $\Lambda_{M_i}^{-1}$ is the updated posterior covariance matrix of the memory after an observation of $\mathbf{z}$.

\subsubsection{Products of Kanerva Machines}

So far, we have dealt only with reformulating the notation for a single Kanerva Machine. What about the product of many Kanerva Machines? We can now consider the full joint distribution between observed $\mathbf{z}$ and all of the memory matrices, which we assume to factor according to the product of the individual joint Gaussian distributions between $\mathbf{z}$ and each memory:
\begin{align}
    p(\mathbf{z}, M_1, \dots M_i, \dots) &\propto \prod_{i=1}^k \, p(\mathbf{z}, M_i) \\
    &\sim \N{\mu, \Lambda^{-1}}
\end{align}

From the mean and precision form of $p(\mathbf{z}, M_i)$, and using the fact that the precision matrix of a product of Gaussians is the sum of the individual precision matrices and that the mean is a precision weighted average of the individual means

\begin{equation}
\Lambda_\text{product} = \Lambda_1 + \Lambda_2
\end{equation}

\begin{equation}
\mu_\text{product} = (\Lambda_1 + \Lambda_2)^{-1}(\Lambda_1\mu_1 + \Lambda_2\mu_2)
\end{equation}

we have the joint precision matrix

\begin{align}
    \Lambda &= \left[ \begin{matrix}
    \Lambda_z & \Lambda^\top_{c_1} & \Lambda^\top_{c_2} & ... \\
    \Lambda_{c_1} & \Lambda_{M_1} & &\\
    \Lambda_{c_2} & & \Lambda_{M_2} &\\
    \vdots &&& \\
    \end{matrix}\right]
        \label{eq:full-joint-cov}
\end{align}
By completing the square, we can compute the parameters of the conditional $p(\mathbf{z} | M_{1:m})$:
\begin{align}
    \Lambda_z &= \sigma_z^{-2} =   \sum_{i=1}^k \sigma_i^{-2} 
\end{align}
and the joint mean
\begin{align}
    \mu &= \left[ \begin{matrix}
    \mathbf{\mu_z} \\
    \vect{R_1}\\
    \vect{R_2}\\
    \vdots
    \end{matrix}\right] \\
    \mathbf{\mu_z} &= \sum_{i=1}^k \gamma_i \, R_i \, \mathbf{w_i}
    \label{eq:cond-mean}
\end{align}
where the coefficient $\gamma_i$ is the normalised accuracy
\begin{equation}
    \gamma_i = \frac{\sigma^{-2}_i}{\sum_{j=1}^k \sigma^{-2}_j}
    \label{eq:alpha-product}
\end{equation}
and $k$ is the number of machines. 

Note that in the block matrix of equation \ref{eq:full-joint-cov}, only the upper left corner couples between the blocks for the different machines/memories. Thus, the posterior update of the covariance, which does not depend on this term, is unmodified compared to the case of individual uncoupled Kanerva Machines. 

The memory update rule for $p(M_i | \mathbf{z})$ is modified as:
\begin{align}
    \Delta &= \mathbf{z} - \mathbf{\mu_z} \\
    R_i &\leftarrow R_i + \beta_i \, \Delta \, \mathbf{w_i^\top} \, V_i \\
    V_i &\leftarrow V_i - \beta_i \, V_i \, \mathbf{w_i}  \mathbf{w_i^\top} \, V_i
\end{align}
where
\begin{align}
     \beta_i &= \frac{1}{\mathbf{w_i^\top} \, V_i \, \mathbf{w_i} + \sigma^2_i} \\
     \label{eq:beta-product}
\end{align}
Note that the only thing that makes this different than independent machine updates is the change in the prediction error term $\Delta$ which, which now couples the machines via $\mathbf{\mu_z}$ from Eq.~\ref{eq:cond-mean}.

Readout takes the form of a simple precision weighted average $\mathbf{\mu_z}$ of the outputs of each individual machine, again from Eq.~\ref{eq:cond-mean}. 

\subsubsection{Generalized Products of Kanerva Machines}

Following \cite{cao2014generalized} we further consider a ``generalized product model'' in which each term in the product of joint distributions may be weighted to a variable amount by raising it to a positive power $r_i$, such that
\begin{equation}
    p(\mathbf{z}, M_i| r_i) \propto p(\mathbf{z}, M_i)^{r_i}
\end{equation}

Since a Gaussian raised to a power is equivalent multiplication of the precision matrix by that power, we may simply replace $\Lambda_i \rightarrow \Lambda_i r_i$ in the above derivation of the product model, for each individual Kanerva Machine, and then proceed with the derivation as normal. The readout equations \ref{eq:cond-mean} and \ref{eq:alpha-product} for $\mathbf{\mu_z}$ are
\begin{align}
\gamma_i \rightarrow \frac{r_i / \sigma_i^2}{\sum_{j=1}^{k} r_j / \sigma_j^2}
\end{align}
\begin{align}
\mathbf{\mu_z} &= \sum_{i=1}^k \gamma_i \, R_i \, \mathbf{w_i}
\end{align}

Meanwhile, in the update equations, we replace $V_i \rightarrow V_i / r_i$ and $\sigma^2_i \rightarrow \sigma^2_i / r_i$, leading to: \begin{align}
    \Delta &= \mathbf{z} - \mathbf{\mu_z}  \\
    R_i &\leftarrow R_i + \beta_i \, \Delta \, \mathbf{w_i^\top} \, V_i \\
    V_i &\leftarrow V_i - \beta_i \, V_i \, \mathbf{w_i}  \mathbf{w_i^\top} \, V_i
\end{align}
where
\begin{align}
     \beta_i &= \frac{1}{\mathbf{w_i^\top} \, V_i \, \mathbf{w_i} + \sigma^2_i / r_i}
\end{align}
\begin{align}
\mu_z &= \frac{\sum_{i=1}^k \frac{r_i}{\sigma_i^2} \, R_i \, \mathbf{w_i}}{\Sigma_{j=1}^k \frac{r_j}{\sigma_j^2}}
\end{align}

This gives our update equations 2-4 in the main text.

Note that $1/\eta_i^2 \defeq r_i/\sigma_i^2$ may also be treated as a single parameter here, and $\eta_i$ generated as the output of a neural network. $\eta_i$ then serves as an effective observation noise $\sigma_i$ for the posterior update of machine $i$.

\begin{remark}
We can understand the coupling between machines during update by expanding the prediction term in equation~\ref{eq:prod-m-update}:
\begin{equation}
    \Delta = \left(\mathbf{z} - \sum_{j \neq i} \gamma_j R_j \mathbf{w_j} \right) - \gamma_i \, R_i \, \mathbf{w_i}
\end{equation}
where terms in the bracket represent the residual from all other $j \neq i$ machines' predictions. Therefore, machine $i$ is updated to reduce this residual, which may then change the residual for other machines. Because of this inter-dependency, the updates of machines are coupled and may take several iterations to converge.

\textbf{In practice, we use a single iteration, making the model fully parallelizable over the $k$ machines.}
\end{remark}

\subsection{Algorithms for writing and reading}
\label{algorithmdef}

Here we present pseudocode for writing and reading in the Product Kanerva Machine.

For clarity, generation and optimization of the ELBO is treated separately in Supp.~\ref{howtogenerate}.  

\begin{algorithm}[H]
\small
\caption{Generalized Product Writing}
\label{Alg:product_writing}
\begin{algorithmic}
\State {\bf Input:} Input episode $\{{x_t}\}_{t=1}^{T}$, and Kanerva Machines $\{M_i\}_{i=1}^{k}$ with means $R_i$ and column covariances $V_i$, total columns $m$, code size $c$, $k$ machines of size $m_i=m/k$ columns each, and $T$ is the episode length. 
\State {\bf Initialization:} Each Kanerva Machine has a trainable prior mean matrix initialized as $R_i^0 \sim \mathcal{N}(0,I)$ and a diagonal prior column covariance with trainable scale, generated as a truncated unit normal $V_i^0 = \psi I$, where $\psi$ is a trainable variable whose logarithm is initialized to $\ln(1.0)$. 
\State The internal slot weighting $\mathbf{w_i}$ of each machine has a prior $p^0({\mathbf{w_i}}) = \mathcal{N}(0, I)$. ${w_i}$ is sampled from a normal distribution with mean produced by a least squares solution, and a diagonal covariance with standard deviation $\chi$, where $\chi$ is a trainable variable whose logarithm is initialized to $\ln(0.3)$. $\Omega_0$ is initialized to $\vec{0}$.
\For{t = 1,...,$T$}
\State $\mathbf{z_t} \gets e(\mathbf{x_t})$ where $\mathbf{e}$ is the ConvNet encoder with output dim $c$
\State
\State $\mathbf{h}_{t-1} \sim q_\phi(\mathbf{h}_{t-1} | z_{1:t-1}) = N(\mathbf{\mu}, \mathbf{\sigma})$ where $\mathbf{\mu} = \text{MLP}_a(\Omega_{t-1})$ and $\mathbf{\sigma} = \text{MLP}_b(\Omega_{t-1})$.
\State Define $\N{\mathbf{\mu}_r, \mathbf{\sigma}_r^2}$ with $\mathbf{\mu_r} = \text{Linear}_1([\text{MLP}_1(\mathbf{z_t}), \mathbf{h}_{t-1}])$, $\mathbf{\sigma}_r = \text{Linear}_2([\text{MLP}_1(\mathbf{z_r}), \mathbf{h}_{t-1}])$
\State $\eta \sim \N{\mathbf{\mu}_r, \mathbf{\sigma}_r^2}$ with $\eta_i$ representing $\sigma_i/\sqrt{r_i}$
\State $\mathbf{\gamma}$: $\gamma_i = \frac{1}{\eta_i^2} / \sum_{j=1}^k \frac{1}{\eta_i^2}$ representing $\gamma_i \rightarrow \frac{r_i / \sigma_i^2}{\sum_{j=1}^{k} r_j / \sigma_j^2}$
\For{j=1,..., $k$}: 
\State $\mathbf{w_j} \gets \text{LeastSquaresOptimize}(M_j, \mathbf{z_t})$ 

\EndFor
\State $\Delta \gets \mathbf{z_t} - \sum_{l=1}^k \gamma_l R_l \mathbf{w_l}$
\For{i=1, ..., $k$}
\State $V_i \gets V_i - \frac{1}{\mathbf{w_i^\top} \, V_i \, \mathbf{w_i} + \eta_i^2}\, V_i \, \mathbf{w_i} \, \mathbf{w_i^\top} \, V_i$
\State $R_i \gets R_i +  \, \Delta \frac{1}{\mathbf{w_i^\top} \, V_i \, \mathbf{w_i} + \eta_i^2} \mathbf{w_i^\top} V_i$
\EndFor
\State $\Omega_t \gets \frac{1}{t}\text{Linear}_0([\mathbf{z_t}, \mathbf{\gamma}]) + \frac{t-1}{t}\Omega_{t-1}$ 
\EndFor
\end{algorithmic}
\end{algorithm}

\begin{algorithm}[H]
\small
\caption{Generalized Product Reading}\label{Alg:product_reading}
\begin{algorithmic}
\State {\bf Input:} Input episode $\{\mathbf{x_t}\}_{t=1}^{T}$, and filled Kanerva Machines $\{M_i\}_{i=1}^{k}$ and history variable $\mathbf{h}_0$.
\For{t = 1,...,$T$}
\State $\mathbf{z}_t \gets \mathbf{e(\mathbf{x_t})}$ where $e$ is the ConvNet encoder with output dim $c$
\State Define $\N{\mathbf{\mu}_r, \mathbf{\sigma}_r^2}$ with $\mathbf{\mu_r} = \text{Linear}_1([\text{MLP}_1(\mathbf{z}_t), \mathbf{h}_{t-1}])$, $\mathbf{\sigma}_r = \text{Linear}_2([\text{MLP}_1(\mathbf{z}_t), \mathbf{h}_{t-1}])$
\State $\eta \sim \N{\mathbf{\mu}_r, \mathbf{\sigma}_r^2}$ with $\eta_i$ representing $\sigma_i/\sqrt{r_i}$
\State $\mathbf{\gamma}$: $\gamma_i = \frac{1}{\eta_i^2} / \sum_{j=1}^k \frac{1}{\eta_i^2}$ representing $\gamma_i \rightarrow \frac{r_i / \sigma_i^2}{\sum_{j=1}^{k} r_j / \sigma_j^2}$
\For{j=1,...,$k$}
\State $\mathbf{w_j} \gets \text{LeastSquaresOptimize}(M_j, \mathbf{z_t})$
\EndFor
\State $\mathbf{\mu_z} \gets \sum_{j=1}^k \gamma_j R_j \mathbf{w_j}$
\EndFor
\end{algorithmic}
\end{algorithm}

\subsection{Mixture model}
\label{mixturemodeldef}

A mixture Kanerva Machine model has $k$ mixture coefficients $\gamma_i$, forming a categorical distribution. The categorical distribution is sampled to yield a one-hot vector $\hat{\gamma_i}$. We then have a read output
\begin{align}\mathbf{z_\text{read}} \gets \sum_{i=1}^k \hat{\gamma_i} R_i \mathbf{w_i}\end{align}
and the writing update for machine $i$ is
\begin{align}
V_i \gets V_i - \hat{\gamma_i} \frac{1}{\mathbf{w_i^\top} \, V_i \, \mathbf{w_i} + \sigma_i^2}\, V_i \, \mathbf{w_i} \, \mathbf{w_i^\top} \, V_i \\
R_i \gets R_i +  \, \hat{\gamma_i} (\mathbf{z_t} - R_i \, \mathbf{w_i}) \frac{1}{\mathbf{w_i^\top} \, V_i \, \mathbf{w_i} + \sigma_i^2} \mathbf{w_i^\top} V_i
\end{align}

The generalized product model becomes a mixture model when $\mathbf{r}$ is one-hot. To see this, note that in this case, in the Product Kanerva Machine, the prediction error $\Delta$ for the single machine $i$ for which $r_i = 1$ inside the product becomes $z-R_i\mathbf{w}_i$ while the readout simply becomes $R_i\mathbf{w}_i$, as in a single Kanerva Machine, while in writing we reduce to the formula for $\beta$ for a single Kanerva Machine. If $i$ is such that $r_i = 0$ we have no readout from that machine and $\beta$ in writing becomes 0 since the denominator becomes $\infty$. Thus, choosing $\mathbf{r}$ as one-hot thus corresponds to selecting a single machine while ignoring the others, while a mixture model corresponds to a stochastic choice of such a one-hot $r$.

We trained such a mixture mixture model using categorical reparametrization via Gumbel-SoftMax sampling~\citep{jang2016categorical, maddison2016concrete} of the machine choice variable. We verified that the Gumbel-SoftMax procedure was resulting in gradient flow using stop-gradient controls.

The mixture model shows MNIST digit class selective machine usage (Fig. 4A-C), but its performance degraded (Fig. 4D) as a fixed total number of slots was divided among an increasing number of machines $k$, in contrast to the robust performance of the product model in Fig. 1 of the main text. 

Note that in the RGB binding task (Fig. 2), the network spontaneously found weights approaching $\{0,1\}$, but it was able to explore a continuous space of soft weights in order to do so, unlike in a mixture model where the weights are one-hot once sampled.

\begin{figure}[h]
\begin{center}
\includegraphics[scale=0.56]{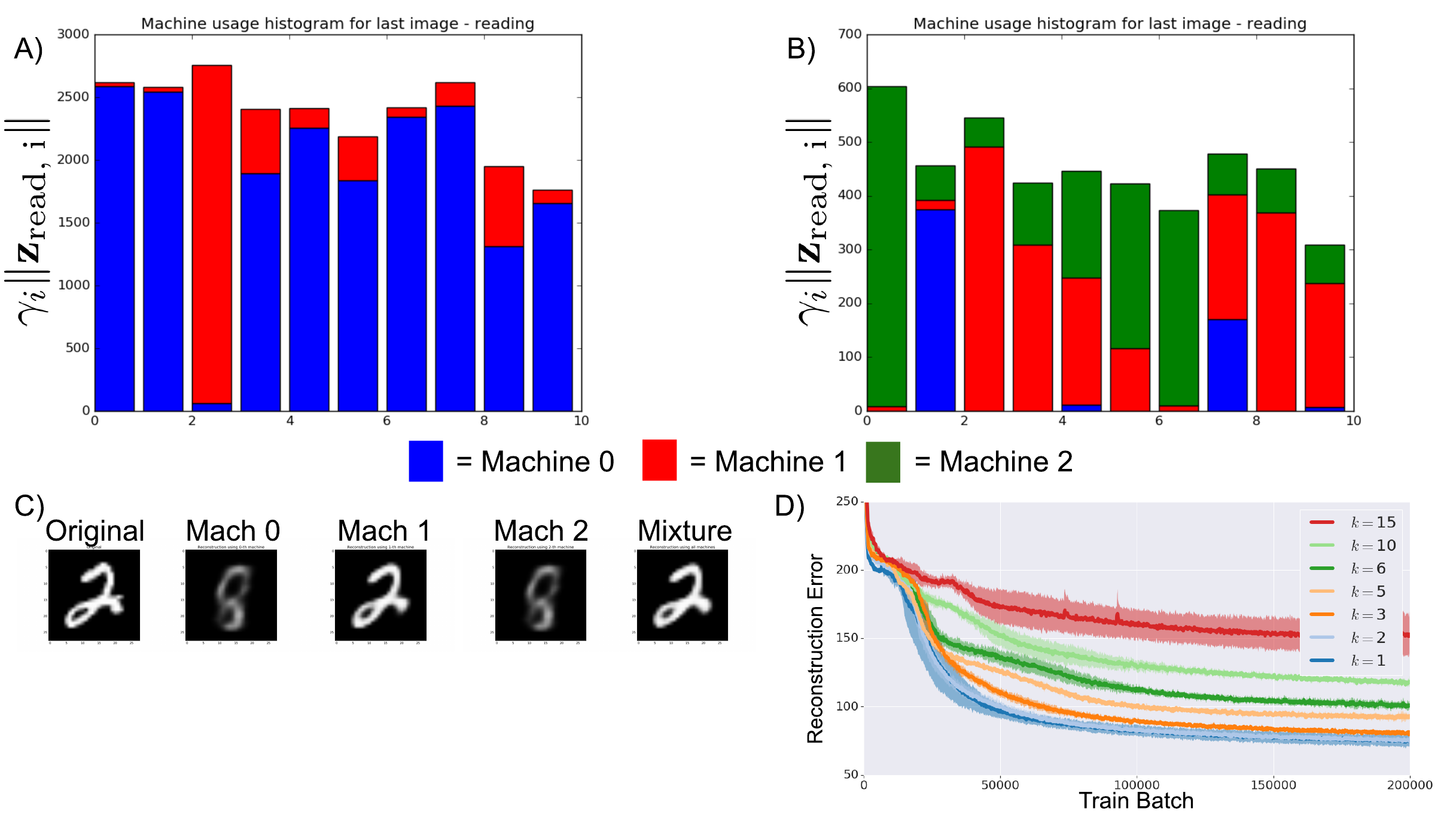}
\end{center}
\caption{Mixture model result on queried MNIST reconstruction. A) Machine usage (norm of vector read from machine $i$ during reconstruction, times the read weight for machine $i$) as a function of digit class for $k=2$ machines. B) Machine usage as a function of MNIST digit class (0-9) for $k=3$ machines. C) Queried single machine reconstructions, and full mixture reconstruction (far right) for a mixture model with 3 machines. Machines $0$ and $2$ are displaying a degenerate class-agnostic pattern corresponding to the average of all MNIST digits, while machine $1$ is responsible for reconstructing this digit, consistent with its dominance for digit class $2$ in panel B. D) Training curve for a mixture model at fixed total slots $m=30$ and increasing $k$, showing performance degradation with $k$.}
\end{figure}

\subsection{Full RGB binding task selectivity matrix}
\label{allrgbclasses}

\begin{figure}[h]
\begin{center}
\includegraphics[scale=0.7]{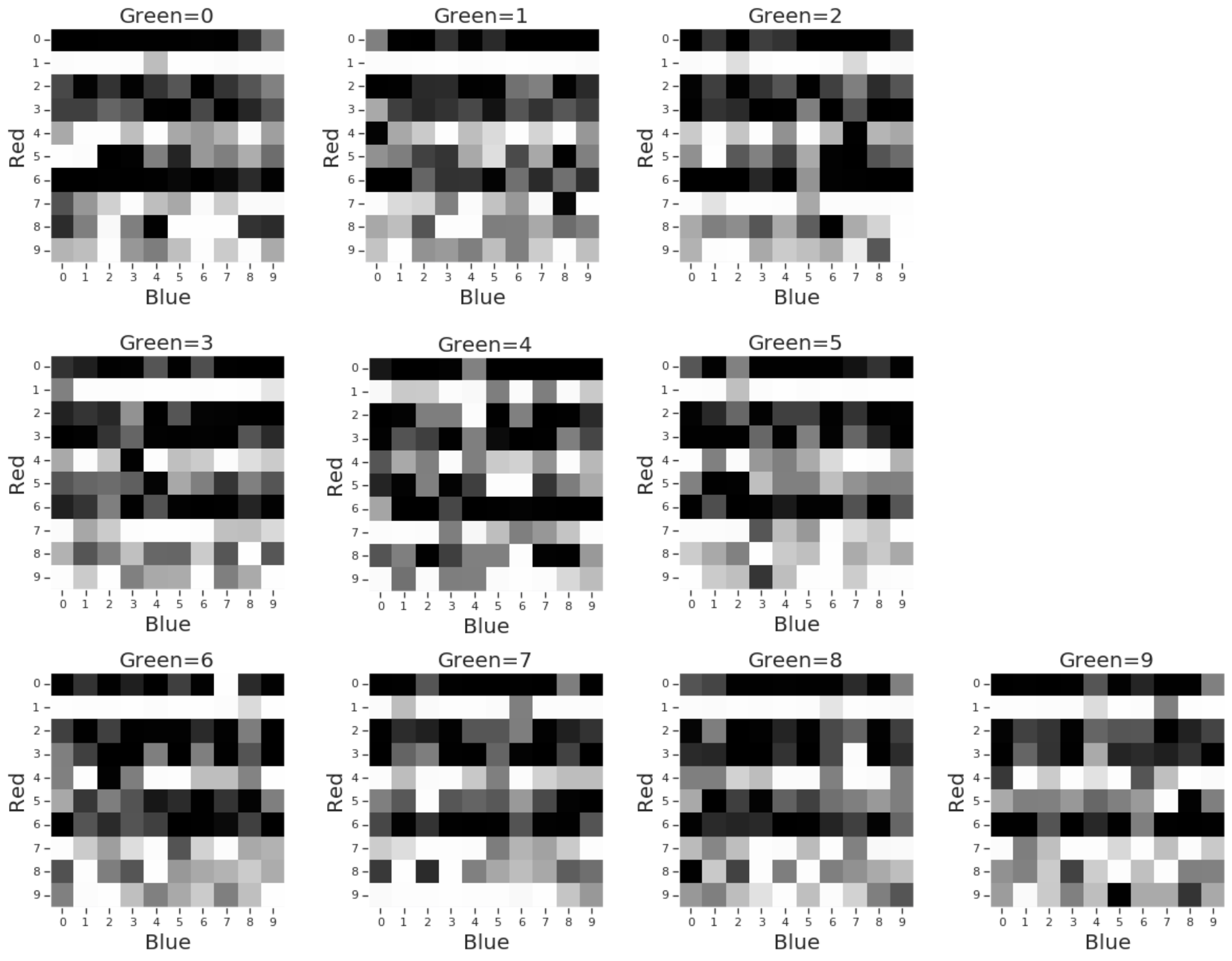}
\end{center}
\caption{Full selectivity matrix for MNIST RBG binding task with a $k=2$ product model. In this example, machine choice is sensitive to the digit class of the Red channel but relatively insensitive to that of the Blue or Green channels. Y: red digit class. X: blue digit class. Image: green digit class. Grayscale within each image: the machine assignment weight $\gamma_1$.}
\label{fig:RGBbindingfull}
\end{figure}

Fig.~\ref{fig:RGBbindingfull} shows the full machine usage matrix for $k=2$ machines on the RGB binding task as a function of the R, G and B MNIST digit classes.

\subsection{Additional representative reconstructions from dancing dSprites task}
\label{sup-objectfactor}

\begin{figure}[h]
\begin{center}
\includegraphics[scale=1.37]{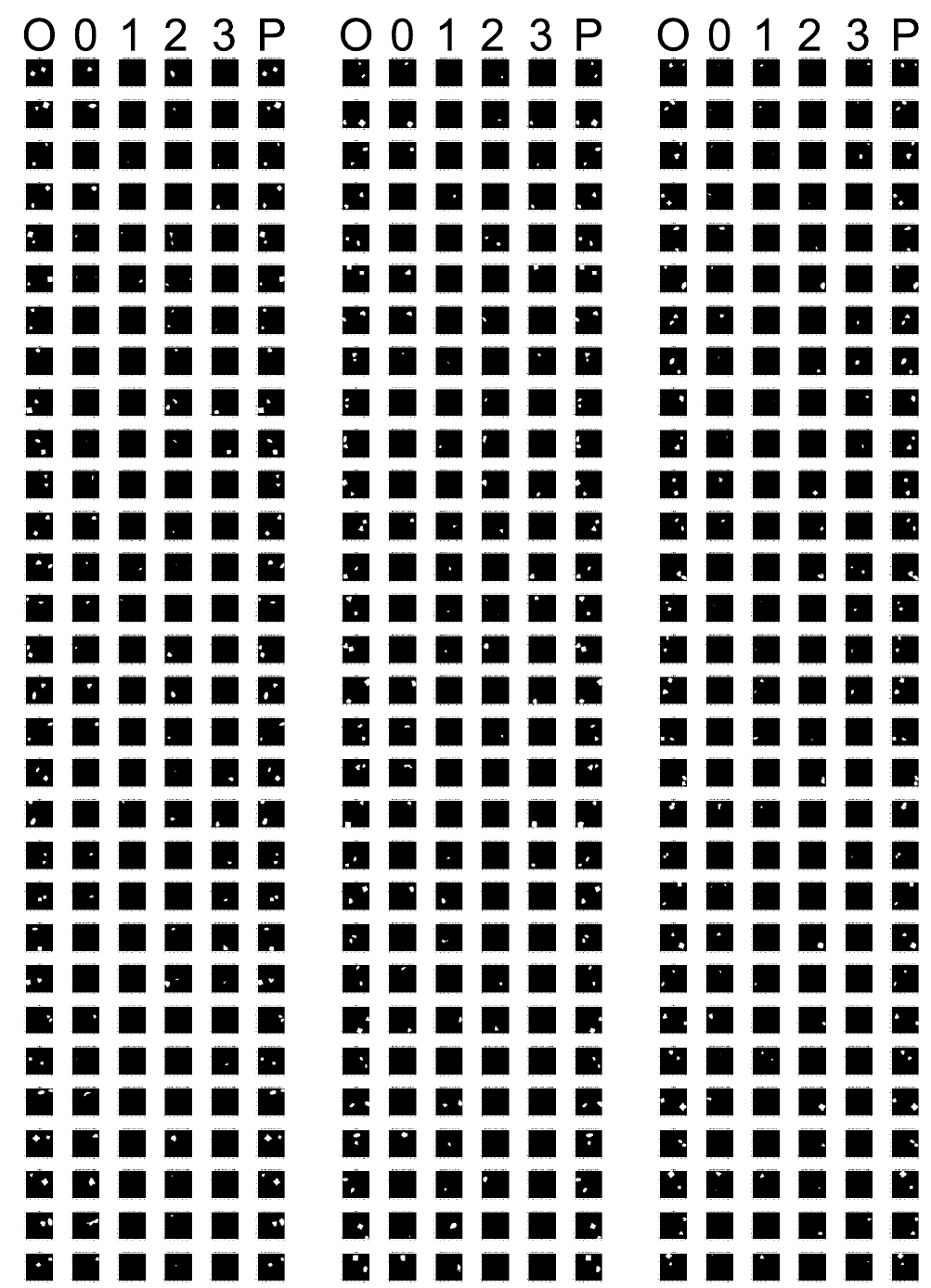}
\end{center}
\caption{Additional dancing dSprite reconstructions showing object separation and spatial tuning. Four individual runs are shown (columns), with many examples from each run (rows). For each run, reconstructions were taken after each of 30 successive 5000 train steps (rows) starting at 100000 train batches. O: Original. 0-3: Reconstructions from each single machine when queried with the full image. P: Product reconstruction.}
\label{fig:additional_dsprite_fig}
\end{figure}

Fig.~\ref{fig:additional_dsprite_fig} shows reconstructions from four different training runs on the dancing dSprites task with $k=4$ machines, episode length $T=15$ and $m=20$ total columns.

\subsection{Dancing dSprite selectivities and invariances}
\label{supp:otherinvariances}

\begin{figure}[h]
\begin{center}
\includegraphics[scale=0.55]{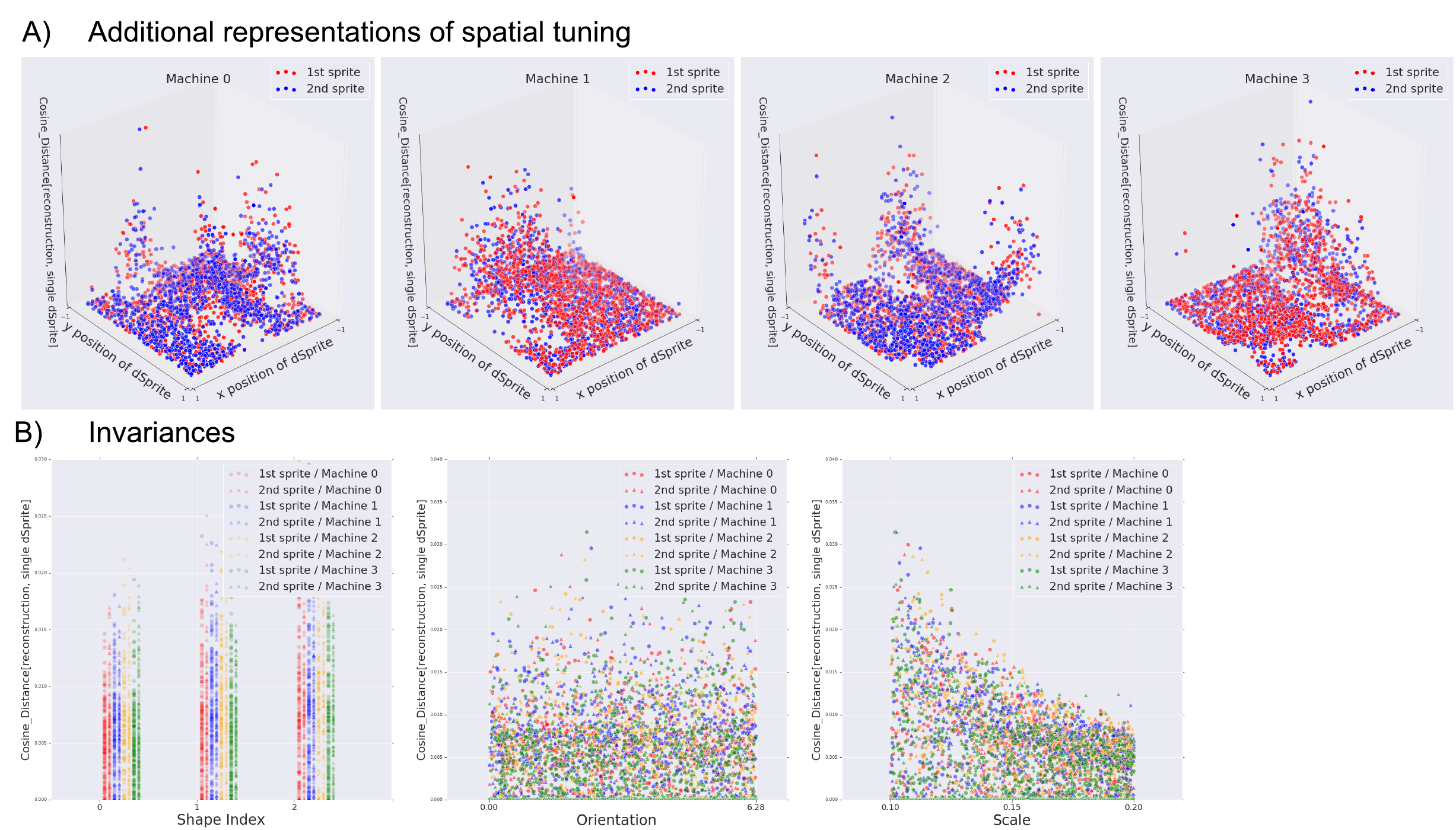}
\end{center}
\caption{Additional selectivities and invariances in the dancing dSprites task. A) Spatial tunings of each of the $k=4$ machines to each of the 2 dSprite positions. Note how tunings across the four machines approximately tile space. B) Tunings to other properties of the individual dSprites: shape (oval, square or heart), orientation ($0$ to $2\pi$) and scale. These properties are invariant across machines. Note that the slope in the curves with respect to scale (right) is due to the fact that single machine reconstructions are typically smaller than the template dSprites they are matched to.}
\label{fig:additionaselectivity}
\end{figure}

Tunings of individual machine reconstructions to dSprite properties were spatially localized and diverse, and seemed to approximately uniformly tile space across the machines, with machines $0$ and $2$ responsible for edges (Fig.~\ref{fig:additionaselectivity}A), but were invariant to shape, orientation and size (Fig.~\ref{fig:additionaselectivity}B). The slope of the curve with respect to size is an artifact of the template matching procedure and the fact that single machine reconstructions are typically smaller than the template dSprites they are matched to.

\subsection{Conditional generations}
\label{sup-condgen}

\begin{figure}[h]
\begin{center}
\includegraphics[scale=0.55]{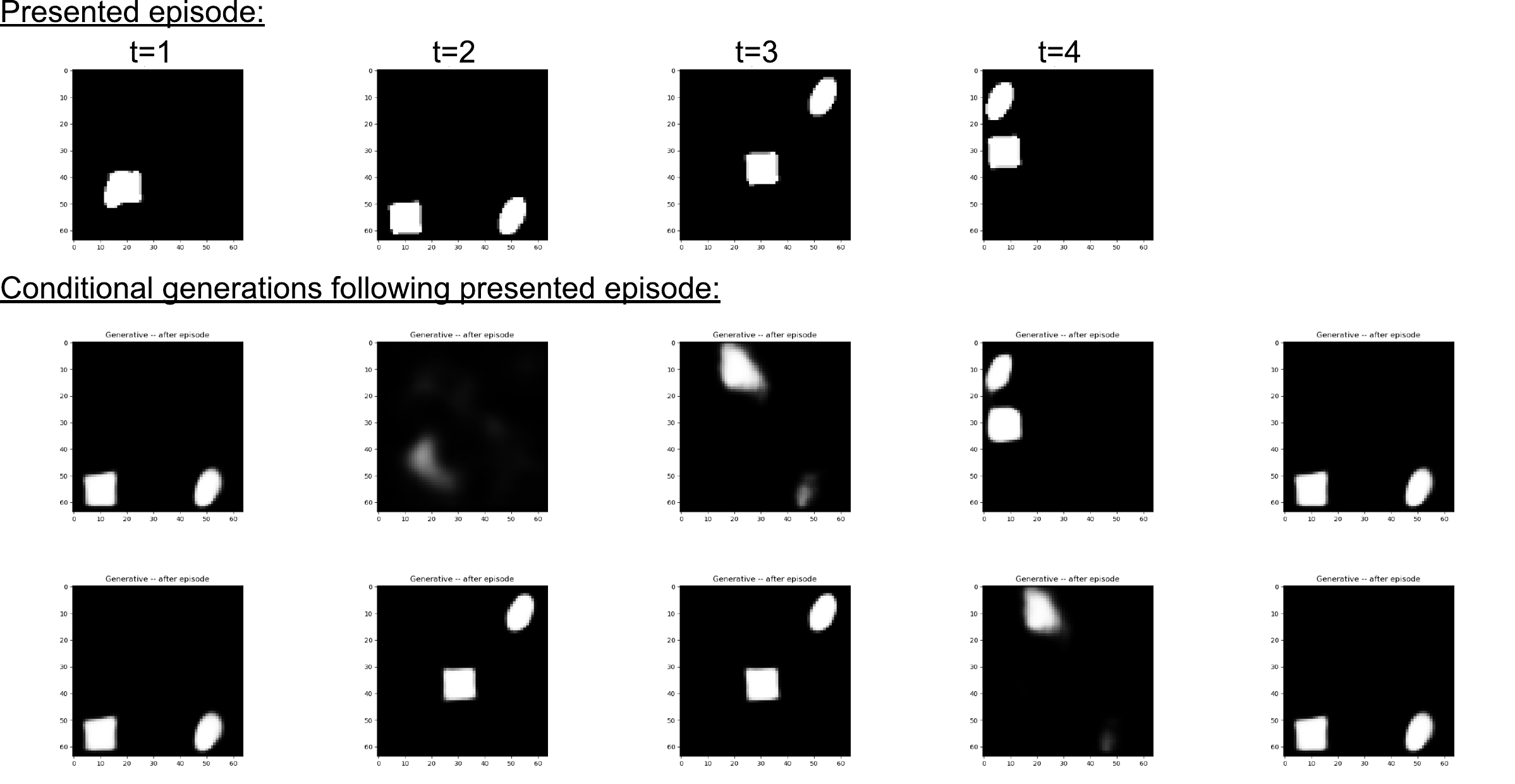}
\end{center}
\caption{Conditional generation example with $k=4$. The top row of images (``presented episode'') was loaded into memory, and then 10 generative samples were taken (bottom two rows) without further updates to the memory. 12 iterations of attractor settling~\citep{dkm2018} were used to generate each image.}
\label{fig:condgen}
\end{figure}

Example conditional generations from the Product Kanerva Machine with $k=4$ and $m_i=5$ after loading a short episode of four dancing dSprite images (Fig.~\ref{fig:condgen}). Twelve iterations of ``attractor settling'' were used~\citep{dkm2018}. In several of the generations the memory has simply retrieved a stored item, whereas in a few generations the model hallucinates noisy spatially localized patterns.

\end{document}